\documentclass[graybox]{styles/svmult}

\usepackage{mathptmx}       
\usepackage{helvet}         
\usepackage{courier}        
\usepackage{type1cm}        
\usepackage{graphicx}
\usepackage{makeidx}         
\usepackage{graphicx}        
\usepackage{multicol}        
\usepackage[bottom]{footmisc}
\usepackage{amsmath}

\makeindex             

\definecolor{mygray}{rgb}{.3,.3,.3}

\begin{document}

\title*{Visual Analytics of Image-Centric Cohort Studies in Epidemiology}
\author{Bernhard Preim, Paul Klemm, Helwig Hauser, Katrin Hegenscheid, Steffen Oeltze, Klaus Toennies, and Henry V{\"o}lzke}
\institute{Bernhard Preim \at Otto-von-Guericke University
Magdeburg, 39106 Magdeburg, \email{bernhard.preim@ovgu.de}}
%
%
\maketitle \abstract{Epidemiology characterizes the influence of
causes to disease and health conditions of defined populations.
Cohort studies are population-based studies involving usually
large numbers of randomly selected individuals and comprising
numerous attributes, ranging from self-reported interview data to
results from various medical examinations, e.g., blood and urine
samples. Since recently, medical imaging has been used as an
additional instrument to assess risk factors and potential
prognostic information. In this chapter, we discuss such studies
and how the evaluation may benefit from visual analytics. Cluster
analysis to define groups, reliable image analysis of organs in
medical imaging data and shape space exploration to characterize
anatomical shapes are among the visual analytics tools that may
enable epidemiologists to fully exploit the potential of their
huge and complex data. To gain acceptance, visual analytics tools
need to complement more classical epidemiologic tools, primarily
hypothesis-driven statistical analysis.}

\section{Introduction}
\label{Introduction}
Epidemiology is a scientific discipline that provides reliable
knowledge for clinical medicine focusing on prevention, diagnosis
and treatment of diseases \cite{Fletcher2011}. Research in
epidemiology aims at characterizing \emph{risk factors} for the
outbreak of diseases and at evaluating the efficiency of certain
treatment strategies, e.g., to compare a new treatment with an
established gold standard. This research is strongly
hypothesis-driven and statistical analysis is the major tool for
epidemiologists so far. Correlations between genetic factors,
environmental factors, life style-related parameters, age and
diseases are analyzed. The data are acquired by a mixture of
interviews (self-reported data, e.g., about nutrition and previous
infections) and clinical examinations, such as measurement of
blood pressure. Statistical correlations, even if they are strong,
may be misleading because they do not represent \emph{causal
relations}. As an example, the slightly reduced risk of heart
infarct and cardiac mortality for elderly people reporting to
drink one glass of wine every evening (compared to people drinking
no alcohol at all) may be due to the involved low level of alcohol
but may also be a consequence of a very regular and stress-free
lifestyle \cite{Fletcher2011}. When something happened,
\emph{before} an event, it is an indicator for a causal
relationship. However, care is necessary, since many things happen
in the life of an individual before, e.g., a heart attack, but do
\emph{not} cause it.

Thus, statistical correlations are the starting point for
investigating \emph{why} certain factors increase the risk of
getting diseases. Epidemiology is not a purely academic endeavor
but has huge consequences for establishing and evaluating
preventive measures even outside of medicine. The protection of
people from passive smoking, recommendations for various
vaccinations and the introduction of early cancer detection
strategies, e.g., mammography screening, are all based on
large-scale epidemiological studies. Also the official guidelines
for the treatment of widespread diseases, such as diabetes, are
based on \emph{evidence} from epidemiological studies
\cite{Fletcher2011}. While this all may sound obvious, it is a
rather recent development. \emph{Evidence-based} medicine often
still has to ``fight" against recommendations of a few opinion
leaders arguing based on their personal experience only.

The analysis techniques used so far are limited to investigating
hypotheses based on known or suspected relations, e.g. hypotheses
related to observations or previous publications. The available
tools support the analysis of a few dimensions, but not of the
hundreds of attributes acquired per individual in a cohort study.
Both typical visualization techniques as well as analysis
techniques, e.g., support vector machines, do not scale well for
hundreds of attributes \cite{Turkay:2013}. While we are not able
to describe \emph{solutions} for these challenging problems, we
give a survey on recent approaches aiming also at \emph{hypothesis
generation}.\newline

\textbf{Organization.} This chapter is organized as follows. In
Sect.~\ref{BackgroundInEpidemiology} we describe important
concepts and terms of epidemiology including observations from
epidemiologic workflows. This discussion is restricted to those
terms that are crucial for communicating with epidemiologists,
understanding requirements and for designing solutions that fit in
their process. In Sect.~\ref{va_epidemiology}, we discuss how
(general) information visualization and data analysis techniques
may be used for epidemiologic data.
Section~\ref{imageDataInEpidemiology} describes the analysis of
image data from cohort studies and how this analysis is combined
with the exploration of non-image attribute data. This section
represents the core of the chapter and employs a case study where
MRI data of the lumbar spine are analyzed along with attributes
characterizing life-style, working habits, and back pain history.

\section{Background in Epidemiology}
\label{BackgroundInEpidemiology}

\textbf{Population-based studies.} Epidemiological studies are
based on a \emph{sample} of the population. The reliability of the
results obviously depends on the size of that sample but also
strongly on the selection criteria. Often, data from patients
treated in one hospital are analyzed. While this may be a large
number of patients, the selection may be heavily biased, e.g.,
since the hospital is highly specialized and diseases are often
more severe or in a later stage compared to the general
population.

\emph{Population-based} studies, where representative portions of
a population (without known diseases) are examined, have the
potential to yield highly reliable results. The source population
may be from a city, a region or a country. Individuals are
randomly selected, e.g., approaching data bases of population
registries. The higher the percentage of people who accept the
invitation and actually take part in the study, the more reliable
the results are.

In this chapter, we focus on longitudinal population-based
studies. The sheer amount and diversity in terms of type of data
makes it difficult to fully identify and analyze interesting
relations. We will show that information visualization and visual
analytics techniques may provide substantial support that
complements the statistical tools with their rather simple
statistical graphics. Most epidemiological studies were restricted
to nominal (often called categorical) and scalar data, e.g.,
related to alcohol consumption, and body mass index as one measure
of obesity.\newline

\textbf{Image-centric epidemiological studies.} More recently, for
example, in the Rotterdam study \cite{Hofman2009}, also
non-invasive imaging data, primarily ultrasound and MRI data, are
employed. Petersen and colleagues \cite{Peterson2013} report on
six studies involving cardiac MRI from at least 1000 individuals
in population-based studies. These high-dimensional data enable to
answer analysis questions, e.g., how does the shape of the spine
changes as a consequence of age, life style and diseases? We focus
on such \emph{image-centric} epidemiological studies.\newline

\textbf{Epidemiology and public health.} There are different
branches of epidemiology. One branch deals with predictions to
inform public health activities. These include measures in case of
an epidemic -- an acute \emph{public health} problem, mostly
related to infectious diseases. The recent article "computational
epidemiology" \cite{Marathe2013} was focussed on this branch of
epidemiology. Another branch of epidemiology aims at long-term
studies and at findings primarily essential for prevention.
Image-centric cohort studies, the focus of this article, belong to
this second branch. The target user group consists of
epidemiologists who can be expected to have a high level of
expertise in statistics. Thus, their findings involve statistical
significance, confidence intervals and other measures of
statistical power.\newline

\textbf{Healthy aging and pathologic changes.} An essential
problem in the daily clinical routine is the discrimination
between healthy age-related modifications (that may not be
reversed by treatment) and early stage diseases (that may benefit
from immediate treatment). As a consequence, elderly people are
often not adequately treated. As a general goal for
epidemiological studies, better and more reliable markers for
early stage diseases are searched for. The cardiovascular branch
of the Rotterdam study, for example, aims at an understanding of
atherosclerosis, coronary heart disease and ``cardiovascular
conditions at older age" \cite{Hofman2009}.\newline

\textbf{Modern epidemiology.} Epidemiology faces new challenges
due to the rapid progress, e.g., in genetics and sequencing
technology as well as medical imaging. Acquisition of health data
thus becomes cheaper and more precise. In cohort studies, as much
potentially relevant data as possible are acquired as a basis for
an as broad as possible spectrum of analysis questions. This
includes blood, urine and tissue samples, information about
environmental conditions and the social milieu.\newline

\textbf{Visual analytics for modern epidemiology.} In the past,
epidemiology primarily dealt with hypotheses aiming to prove them,
e.g., the efficiency of early cancer detection programs in terms
of mortality and long term survival \cite{Fletcher2011}. Since
recently, more and more data mining is performed to identify
correlations. Results of such analyses, however, need to be very
carefully interpreted. If thousands of potential correlations are
analyzed automatically, just by chance some of them will reach a
high level of statistical significance.

An essential support for epidemiology research is to define
relevant subgroups. To perform separate analyses for women and men
as well as for different age groups is a common practice in
epidemiology. However, relevant subgroups may be defined by a
non-obvious combination of several attributes that may be detected
by a combination of cluster analysis and appropriate
visualization.

Since the information space is growing with each examination
cycle, Pearce and Merletti \cite{Pearce2006} pointed out in 2006
that methods are needed which can cope with this complexity and
enable the analysis of underlying causes of a certain disease.
Visual analytics (VA) methods can support epidemiological data
assessment in different ways, e.g. by defining subgroups based on
a multitude of attributes that exhibit a certain characteristic.
For the analysis of scalar and categorical data, established
information visualization techniques combined with clustering and
dimension reduction are a good starting point, but need to be
tightly integrated with statistic tools epidemiologists that are
more familiar with. For image-centric studies, however, new
visualization, (image) analysis and interaction techniques are
needed.

In the following, we define essential terms in epidemiology and
give an overview on cohort studies that employ medical image data
as an essential element. Finally, we describe how image data,
derived information and other data complement each other to
identify and characterize risks.

\subsection{Important Terms}

\textbf{Prevalence and incidence.} Epidemiology investigates how
often certain diseases or \emph{clinical events}, such as a
cerebral stroke or sudden heart death, occur in the population.
Two terms are important to characterize this frequency. The
\emph{prevalence} indicates the portion of people suffering from a
disease at a given point in time. The \emph{incidence} represents
how many people suffer from a disease or event in a certain
interval, usually one year. High prevalence is usually associated
with high economic costs. Population-based studies focus on
diseases with a high prevalence, such as diabetes, coronary heart
disease or neurodegenerative diseases. Even these diseases do not
occur frequently in a random population including many younger
people (where the prevalence of these diseases is low). A rare
disease, such as amyotrophic lateral sclerosis, may have a
prevalence of 5 from 100,000. Thus, even in a large
population-based study probably no individual suffers from this
disease.\newline

\textbf{Absolute and relative risks.} Another essential
epidemiological term is the \emph{risk} for a clinical event, such
as outbreak of a certain disease, severity (stage) or death. As an
example, a study related to cardiac risk may investigate angina
pectoris, myocardial infarction, atrial fibrillation depending on
attributes such as age and sex. The \emph{absolute risk}
characterizes the likelihood of getting a disease in life time.
The absolute risk for a woman to develop breast cancer in the
Western world is particularly high for women aged 50-60 (2.6\%)
and 60-70 (3.7\%). Therefore, for these age groups, mammography
screening -- aiming at early detection and thus optimal treatment
-- was introduced.

The \emph{relative risk} (RR) characterizes the increased risk if
an individual is exposed to a certain risk factor, e.g., smoking,
excessive weight, or alcohol abuse. It is based on a comparison
with a control group not exposed to that risk factor. A value of
$RR<1$ represents a factor that protects, e.g., moderate physical
activity. Exciting observations are often the combined effects of
several parameters. A certain factor may be protective for some
people (younger, slim women) and is involved with an increased
risk for others. The combined risk may be significantly smaller or
larger than could be expected from individual factors.

Moreover, relationships are often distinctly non-linear or even
non-monotonic. Dose-response relationships are often non-linear.
RR increases slowly (almost no effect for a small dose) and
increases much faster for higher levels of a dose, e.g., exposure
to toxicity. A typical non-monotonic relation is \emph{U-shaped},
that is both very low and very high instances of an attribute
involve an increased risk, whereas values in between are
associated with a reduced risk. Examples are weight (both very low
and very high weight are associated with an increased risk for
mortality) and sleeping time (both very short and very long
sleepers have an increased risk for developing psychiatric
disorders \cite{Hofman2009}). Such relations cannot be
characterized by a global RR value. Instead, tools are necessary
that support the hypothesis of a U-shaped relation by estimating
their parameters with some kind of best-fit algorithm.

\subsection{Image-Centric Cohort Studies}
\label{subsec:imageData}

\textbf{Image data in epidemiology.} The acquisition of image data
is determined by the available time, by financial resources, by
the epidemiological importance and by ethic considerations.
Epidemiological studies require approval by a local ethics
committee. As a consequence, healthy individuals in a cohort study
should not be exhibited to a risk associated to the examinations
carried out. Thus, MRI should be preferred over X-ray or CT
imaging for its non-radiation nature. Petersen and colleagues
\cite{Peterson2013} explain why cardiac CT is less feasible in a
cohort study and even MR is only used without a contrast agent in
their study due to ethical reasons. MRI data and ultrasound data
are the prevailing modalities in both the SHIP as well as the
Rotterdam study. Unfortunately, MRI and ultrasound data do not
exhibit standardized intensity values (in contrast to CT data).
Moreover, MRI and ultrasound data suffer from inhomogeneities and
various artifacts. Thus, they are more difficult to interpret for
humans and more difficult to analyze with computational means.
These data are used to measure, e.g., the thickness of vessel
walls, the abdominal aorta diameter and plaque vulnerability in
the coronary vessels \cite{Hofman2009}. The intensive use of MRI
in epidemiological research also explains to some extent which
questions are analyzed: MRI is the best modality for the analysis
of brain structures and thus serves to explore early signs of
Parkinson's, Alzheimer's and other neurodegenerative diseases.
Epidemiological research aims at identifying such brain
pathologies in a pre-symptomatic stage. Among the sources for such
investigations are MR Diffusion Tensor Imaging data that enable an
assessment of white matter integrity \cite{Hofman2009}.

The selection of imaging parameters is always a trade-off between
conflicting goals related to quality, e.g., image resolution,
signal-to-noise ratio, patient comfort, e.g., examination time and
associated costs. As a consequence, to shorten overall examination
times in cohort study examinations, not the highest possible
quality is available, i.e., a slice distance of 4~mm is more
typical than 1~mm. A great advantage of MRI is that this method is
very flexible and enables to display different structures in
different sequences, such as T1-, T2- and proton density-weighted
imaging. MRI data in cohort studies often comprise more than ten
different sequences.\newline

\textbf{Standardization in image acquisition.} Due to the rapid
progress in medical imaging, sequences, protocols and even (MR)
scanners are frequently updated in clinical routine (similar to
the update frequency on a computer). These updates would severely
hamper the comparison of imaging results and thus the assessment
of natural changes and disease outbreak. Thus, differences in
acquisition parameters are essential \emph{confounding variables}.
Therefore, for one cohort and examination cycle that may last up
to several years, no updates are allowed. Moreover, all involved
physicians and radiology technicians are carefully instructed to
use the same standardized imaging parameters. This point is even
more important for longitudinal studies with repeated imaging
examinations. Even if MR scanners and protocols are not updated,
the life cycle of MR coils leads to changes of image quality that
need to be monitored and compensated.

\subsection{Examples for Image-Centric Cohort Study Data}
\label{subsec:examples}

In the following, we describe selected comprehensive and on-going
longitudinal cohort studies. Both use a number of (epidemiologic)
\emph{instruments} that are innovative in cohort studies and thus
lead already to a large number of insights documented in hundreds
of (medical) publications. A considerable portion of these
publications employ results from imaging data. However, the full
potential of analyzing organ shapes, textures and spatial
relations quantitatively is not exploited so far.\newline

\textbf{The Rotterdam study.} A prominent example is the
\emph{Rotterdam}
Study\footnote{\url{http://www.erasmus-epidemiology.nl/research/ergo.htm},
accessed: 1/31/2014}, initiated in 1990 in the city of Rotterdam,
in the Netherlands. Similar to later studies, it was motivated by
the demographic change with more and more elderly people suffering
from different diseases and their interactions. After the initial
study involving almost 8,000 men and women, follow-ups at four
points in time were performed---the most recent examinations took
place in the 2009-2011 period. In the later examination cycles,
also new individuals were involved leading to datasets from almost
15,000 patients \cite{Hofman2009}.

The original focus of the Rotterdam Study was on neurological
diseases, but meanwhile it has been extended to other common
diseases including cardiovascular and metabolic diseases.  The
study has an enormous impact on epidemiological and related
medical research, documented in 797 journal publications
registered in the pubmed database (search with keyword ``Rotterdam
Study", January 30, 2014). Among them are predictions for the
future prevalence of heart diseases and many studies on potential
risk factors for neurodegenerative diseases. For a comprehensive
overview of the findings, see \cite{Hofman2009} that summarizes
the findings of more than 240 papers related to the Rotterdam
Study. In a similar way, \cite{Hofman2011} is a significant update
of these findings with more recent data.\newline

\textbf{Norwegian Aging Study.}  A long-term study in Norway
investigates the relations between brain anatomy (as well as brain
function), cognitive function, and genetics in normally aging
people.\footnote{\url{http://org.UiB.no/aldringsprosjektet/},
accessed: 1/31/2014} In total 170 individuals (120 of them
female), aged between 46 and 77 (mean 62), were examined in Bergen
and Oslo in by now three waves (1st wave in 2004/2005, next in
2008/2009, and most recently in 2011/2012)~\cite{Ystad2009}. While
naturally not all of these subjects could be followed through all
three waves, still most of them were subjected to an extensive
combination of
\begin{enumerate}
  \item neuropsychological tests, including tests of the
      intellectual, language (memory), sensory/motor, and
      attention/executive function,
  \item MRI data, including co-registered T1-weighted
      anatomical imaging, diffusion tensor imaging, and --
      from the 2nd wave on -- also resting-state functional
      MRI, as well as
  \item  genotyping (1st wave only)~\cite{YstadPhD}.
\end{enumerate}

The substantially heterogeneous imaging and test data are used to
study aging-related questions about the modern Norwegian
population, for example, how anatomical and functional changes in
the human brain possibly relate to the later development of
Alzheimer and dementia. Important findings include the relation
between hippocampal volumes and memory function in elderly
women~\cite{Ystad2009} and the relation between subcortical
functional connectivity and verbal episodic memory function in
healthy elderly~\cite{Ystad2010}.\newline

\textbf{SHIP.} The Study of Health in Pommerania (SHIP) is another
cohort study broadly investigating findings and their potential
prognostic value for a wide range of diseases. The SHIP tries to
explain health-related differences after the German reunion
between East and West Germany. It was initiated in the extreme
northeast of Germany, a region with high unemployment and a
relatively low life expectancy.

In the first examination cycle (1997-2001) 4,308 adults of all age
groups were examined, followed by a second and a third cycle that
was finished at the end of 2012. The instruments used changed over
time with some initial image data (liver and gallbladder
ultrasound) available already in the first cycle and others, in
particular whole body MRI, added later. The use of whole body MRI
was unique in 2008 when the third examination cycle started.
Breast MRI for women is performed, whereas for men MR angiography
data are acquired, since men suffer from cardiovascular diseases
significantly earlier than women \cite{Volzke2011}. In addition, a
second cohort (SHIP-Trend) was established comprising 4,420 adult
participants.

Diagnostic reports are created by two independent radiologists who
follow strict guidelines to report their findings in a
standardized manner. The pilot study to discuss the viability and
potential of such a comprehensive MR exam is described by
\cite{Hegenscheid2009}. The overall time for the investigation is
two (complete) days with 90 minutes for the MR exam. The SHIP
helped to reliably determine the prevalence of risk factors, such
as obesity, and diseases. Major findings of the SHIP are increased
levels of obesity and high blood pressure (compared to the German
population) in the cohort. The MR exams alone identified
pathological findings in 35\% of the sample population. More than
400 publications in peer-reviewed journals are based on SHIP data
(January 2014).\newline

\textbf{UK Biobank.} The UK Biobank started recently and
represents a comprehensive approach to study diseases with a high
prevalence in an aging society, such as hearing loss, diabetes and
lung diseases. Half a million individuals will be investigated in
one examination cycle from which 100,000 receive an MRI from 2014
onwards. The rationale for the number of individuals to be
included is explained by Peterson and colleagues
\cite{Peterson2013}: they aim at a reliable identification of even
moderate risk factors (RR between 1,3 and 1,5) for diseases with a
prevalence of 5\%. The prospective study should have a
comprehensive protocol of cardiac MRI, brain MRI and abdominal
MRI. This prospective cohort study also involves genetic
information.\footnote{\url{http://www.ukbiobank.ac.uk}, accessed:
1/31/2014}\newline

\textbf{The German National Cohort.} The recently started ``German
National Cohort" in Germany is based on experiences with a number
of moderate-size studies, such as SHIP, and examines some 200,000
individuals over a period of 10-20 years. Individuals will be
invited in three waves to characterize changes. Due to the
large-scale character, imaging is distributed over five cities.
Thus, the subtle differences in imaging within different scanners
have to be considered.
\footnote{\url{http://www.nationale-kohorte.de/}, accessed:
1/31/2014} It explicitly aims at improvements in the treatment of
chronic diseases and involves a variety of tissue samples, e.g.,
lymphocytes. Imaging in 30,000 individuals is again performed with
MRI, comprising whole body, brain and heart.

\subsection{Epidemiological Data}

Epidemiological data are huge and very heterogeneous. As an
example, in the UK biobank 329 attributes relate to physical
measures, such as pulse rate, systolic and diastolic blood
pressure, and various measures relate to vision or hearing. 471
attributes relate to interviews (socio-demographics, health
history, lifestyle, \ldots).

The data that are stored per individual is standardized but not
completely the same, e.g., childbirth status and menstrual period
are available for women only. Image data and derived information,
e.g., segmentation results, significantly increased both the
amount and complexity of data. Longitudinal cohort study data are
time-dependent. While some \emph{instruments}, such as blood
pressure measurements, are available for all examination cycles,
others were added later or removed. Individuals drop out, because
they move, die or just do not accept the invitation to a second or
third examination cycle. It is important to consider also such
incomplete data but to be aware of potentially misleading
conclusions.

The great potential of image-centric studies is that image data
and associated laboratory data as well as data from interviews are
available. An epidemiological study, such as the SHIP, has a large
\emph{data dictionary} that precisely defines all attributes and
their ranges. While laboratory data are scalar values, most data
from interviews are nominal or ordinal values. In particular, data
from interviews exhibit an essential amount of uncertainty.
Self-reports with respect to alcohol and drug use, cigarette
smoking and sexual practices may be biased towards ``expected or
socially accepted" answers. Epidemiologists are not only aware of
these problems but developed strategies to minimize the negative
effects, e.g., by asking redundant questions. After data
collection, experts spend a lot of effort to improve the quality
of the data. Despite these efforts, visual analytics techniques
have to consider outliers, missing and erroneous data.\newline

\textbf{Geographic data.} Geographic data play a central role in
public health where the dynamics of local infections are
visualized and analyzed (\emph{disease mapping}). Chui and
colleagues \cite{Chui:2011gm} presented a visual analytics
solution directly addressing this problem by combining three
dedicated views. Also in cohort study data, geographic data are
potentially interesting to understand local differences in the
frequency and severity of diseases as an interaction between
environmental factors and genetic differences. This branch of
epidemiology is referred to as \emph{spatial epidemiology}. Beale
and colleagues \cite{Beale:2008dz} investigated differences
between rural and urban populations. In their comprehensive
survey, Jerrett and colleagues \cite{Jerrett:2010} considered
spatial epidemiology as an emerging area. However, we do not focus
on spatial epidemiology since cohort study data typically comprise
rather narrow regions and thus may not fully support such analysis
questions.

\subsection{Analysis of Epidemiological Workflow}

The following discussion of observations and requirements for
computer support is largely based on discussions with
epidemiologists as well as the inspiring publication by Thew and
colleagues. According to \cite{Thew:2009vz}
\begin{itemize}
    \item epidemiological hypotheses are mostly observations
        made by physicians in clinical routine,
    \item corresponding attributes are chosen based on the
        observations and further experience, and
    \item regression analysis is frequently used to determine
        whether the investigated attribute is a risk factor
        or not.
\end{itemize}

Major requirements for an epidemiological workflow (again based on
\cite{Thew:2009vz}) are: \vspace{-1.7mm}
    \begin{itemize}
        \item Results have to be reproducible. Due to the
            iterative data assessment, methods need to be
            applied to new data sets as well and the results
            need to be comparable between different assessment
            times to characterize the change. User
        input needs to be monitored all the
            time to enable reproducible results.
        \item A major result of an epidemiological analysis is
            whether certain factors influence a disease
            significantly. Relative risk (as a measure of
            effect size) and p-values as statistical
            significance level are particularly important.
            \vspace{-1.7mm}
    \end{itemize}

Although these requirements neither consider image data nor visual
analytics, they have to be considered also in these more
innovative settings. Reproducibility, for example, means that
clustering with random initialization is not feasible. Moreover,
reports must be generated that clearly reveal all settings, e.g.,
parameters of clustering algorithms that were used for generating
the results.

\begin{figure}[!htb]
\sidecaption
\includegraphics[width=7cm]{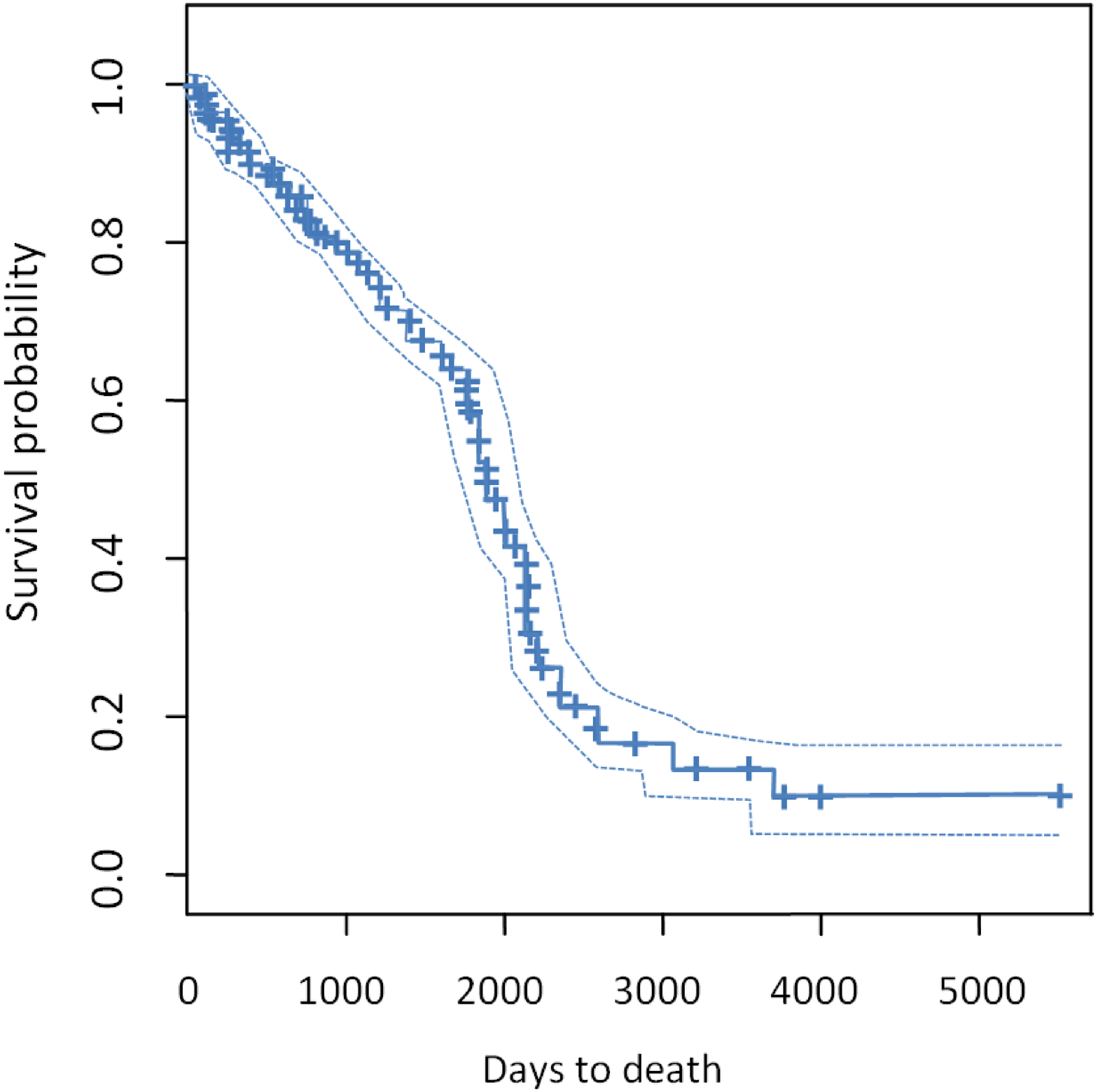}
\caption{A Kaplan-Meier curve indicates how many patients survive at least a certain time. The more patients pass away,
the larger is the confidence interval indicated by the dotted lines. The crosses mark each time a patient dies to further provide
information on the reliability of the data that decreases over time (Courtesy of Petra Specht, University of Magdeburg).}
\label{fig:Kaplan}
\end{figure}

\begin{figure}[!htb]
\sidecaption
\includegraphics[width=7cm]{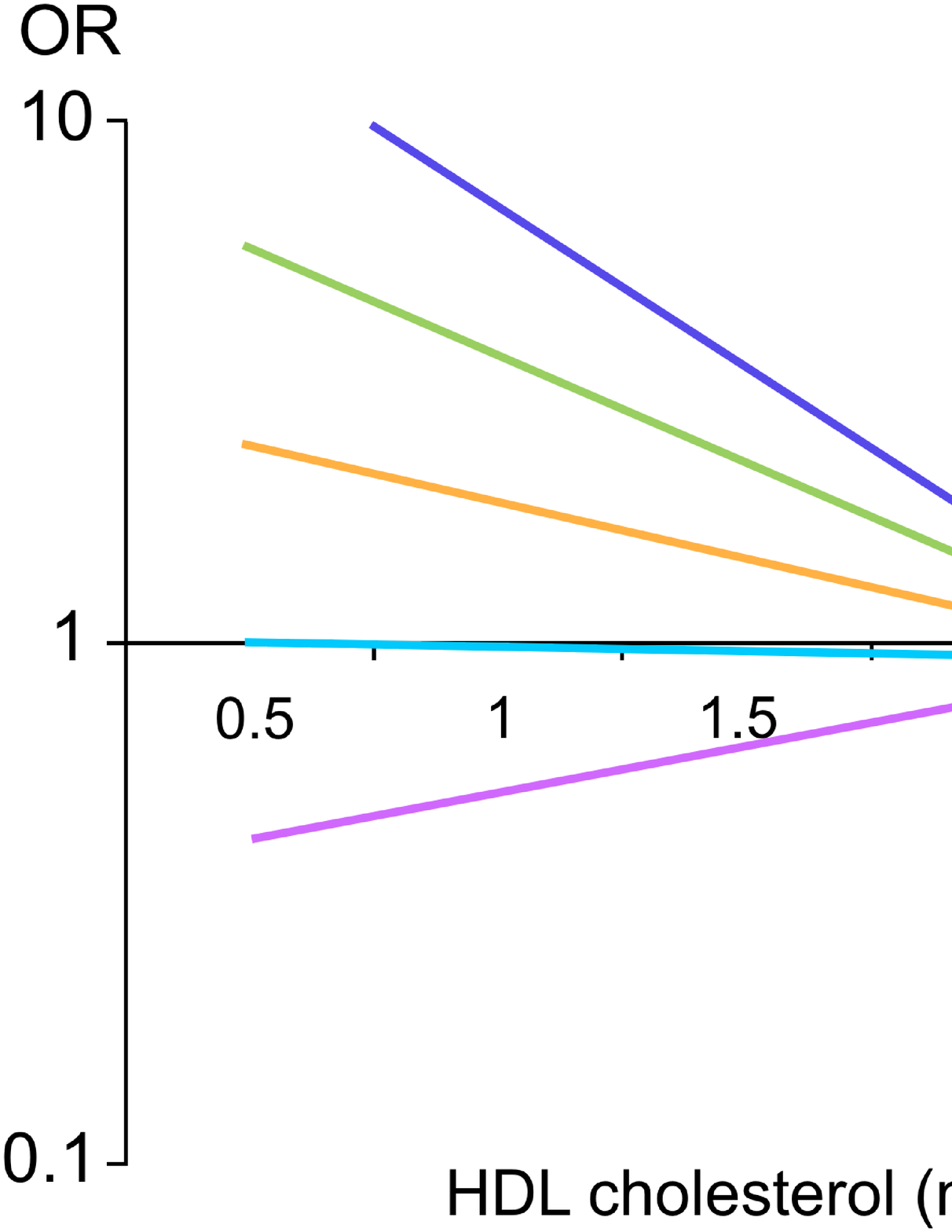}
\caption{The relative risk for cholelithiasis in men associated with a high level of a certain type of cholesterol slightly
increases with a low BMI, but decreases for individuals with high or very high BMI. This multifactorial situation is depicted in an \emph{interaction term}
(Inspired by \cite{Volzke2005}).}
\label{fig:interaction}
\end{figure}

Since statistical analysis plays such an important role,
statistics packages, such as
SPSS\footnote{\url{http://www-01.ibm.com/software/analytics/spss/products/statistics/},
accessed: 1/31/2014}, R\footnote{\url{http://www.r-project.org/},
accessed: 1/31/2014} and
STATA\footnote{\url{http://www.stata.com/}, accessed: 1/31/2014}
dominate in epidemiology. They provide various statistical tests
also in cases where assumptions, such as a normal distribution,
are not valid. Also the peculiarities of categorical data are
considered. Visual analysis, so far, plays a minor role. As an
example, Figures~\ref{fig:Kaplan} and \ref{fig:interaction}
illustrate two graphical representations frequently used in
epidemiology: \emph{Kaplan-Meier curves} and \emph{interaction
terms}. A Kaplan-Meier curve shows the survival of patients, often
as a comparison between different treatment options.

\section{Visual Analytics in Epidemiology}
\label{va_epidemiology}

The visualization of correlations in the epidemiological routine
is largely restricted to scatterplots with regression lines and
box plots to convey a distribution. Scatterplots may be enhanced,
e.g., by coloring items according to certain characteristics,
e.g., a diagnosis or by adding results from cluster or Principal
Component Analysis (PCA) \cite{Steenwijk:2010wo}. The frequency
related to a particular combination of values is often encoded by
adapting saturation or darkness of colors.

Visual analytics methods can complement the statistical analysis
and provide methods to \emph{explore} the data. Efficient methods
are essential to cope with the large amount of data and provide
rapid feedback that is essential for any exploration process.

One of the first attempts to employ information visualization
techniques for medical (image and non-image) data was realized in
the WEAVE system \cite{Gresh:2000}. The system incorporated
parallel coordinates as well as real time synchronization between
different views. Another essential tool, inspired by the WEAVE
system, was presented by Blaas and colleagues \cite{Blaas2007}.
They enabled feature derivation techniques and incorporated
segmentation techniques providing a powerful framework for
heterogeneous medical data. Later work by Steenwijk and colleagues
was more focussed on epidemiology. They provided an exploratory
approach to analyze heterogeneous epidemiological data sets,
including MRI \cite{Steenwijk:2010wo}. They consider parameters on
normalized and not normalized domains, while only normalized
domains are comparable between subjects. Normalization means, for
example, to register MRI brain data to an atlas to compare
individual differences.

\begin{figure}[!htb]
\begin{center}
\includegraphics[height=3.23cm]{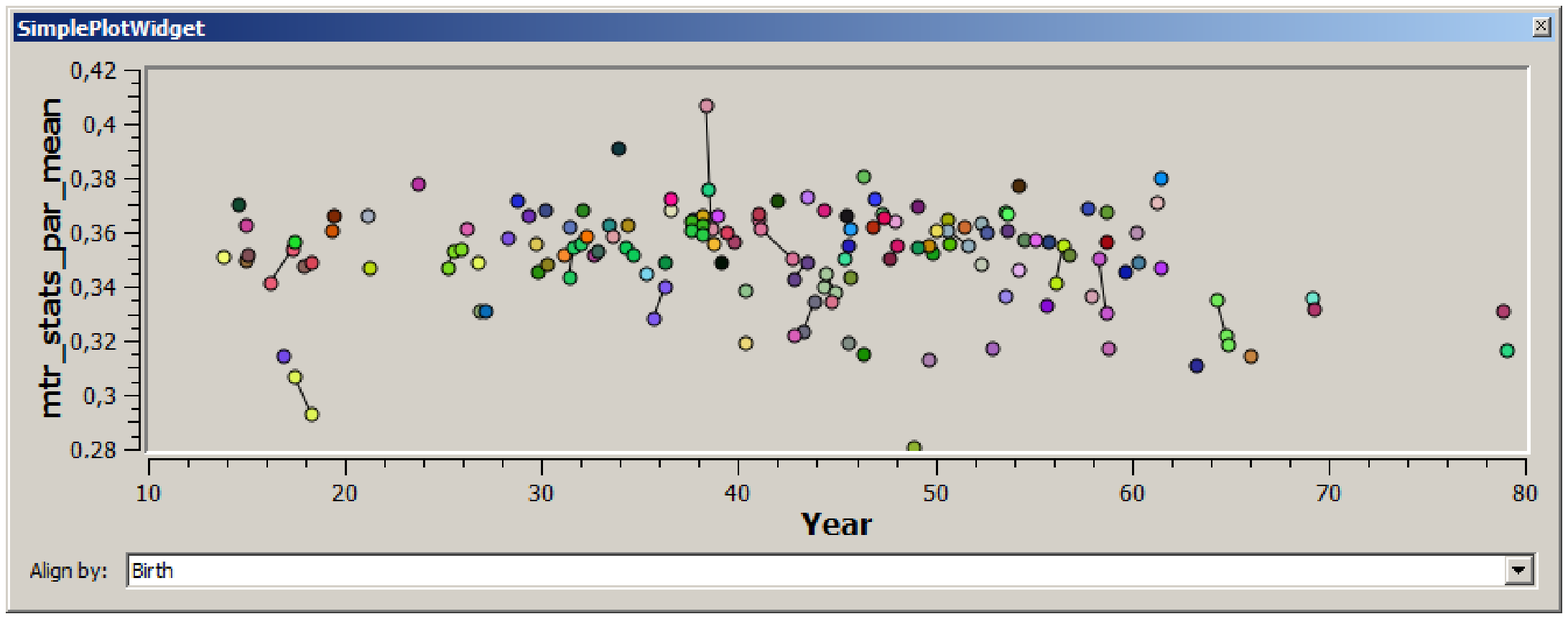}
\includegraphics[height=3.23cm]{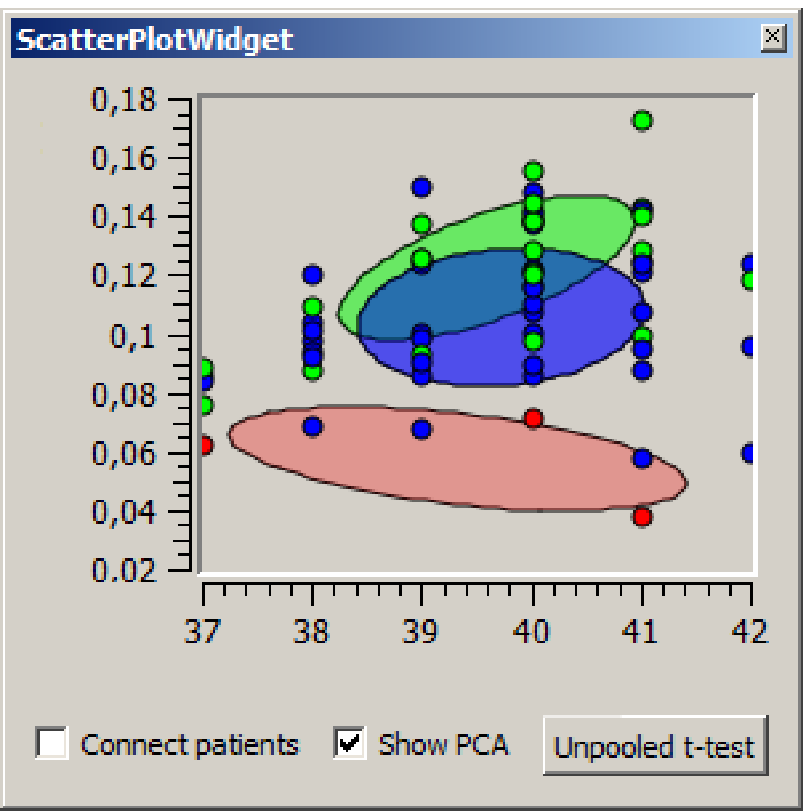}
\end{center}
\caption{Left: A scatterplot relates magnetic transfer ratios to
age. Items relating to the same patient (over time) are connected
via a line. Colors indicate a diagnosis. Right: An enhanced
scatterplot with PCA results for three subgroups overlaid
(Courtesy of Martijn Steenwijk, VU University Medical Center
Amsterdam).} \label{fig:Steenwijk}
\end{figure}

\emph{Mappers} are used to project data into normalized domains.
As an example, in brain analysis, a mapper defines the relation
between an individual brain and a brain atlas that contains
normalized and averaged information derived from many individual
data. Feature extraction pipelines can be build visually by using
a pipeline of mappers. The visualization is realized through
multiple coordinated views which either represent scalar data or
volumetric images. Different techniques to color code data, to
align them and add further information are provided to enhance
scatterplots. Steenwijk and colleagues evaluated their tools with
specific examples from neuroimaging and questions related to a
neurological disease where relations between clinical data
(anxiety-depresssion scales, mental state scales) and MR-related
data are analyzed (Fig.~\ref{fig:Steenwijk}). Normalized data
domains are represented using scatterplots and parallel
coordinated views. Dynamic changes are visualized using a time
plot. The selection is linked between views and allows for
multi-parameter comparison of clusters.

Zhang and colleagues \cite{Zhang:2012uw} build a web-based
information visualization framework for epidemiological analysis
through different views. They divide the analytics process in
\emph{batch analytics} and \emph{on-demand analytics}. Batch
analytics steps are performed automatically as a new subject is
added to the data set and aim to create groups by means of a
certain condition. On-demand analytics are performed by user
requests. Subjects are visualized using treemaps, histograms,
radial visualizations and list views. However, neither filtering
and grouping nor the interaction between the views are explained.

Recently, Turkay and colleagues \cite{Turkay:2013} described a
framework to analyze the data of the Norwegian aging study, aiming
particularly at \emph{hypothesis generation}. For this purpose,
they give an overview on the dimensions in their dataset that
conveys statistical properties, such as mean, standard deviation,
skewness and kurtosis. The two latter measures characterize how
asymmetric the data distribution is. Scatterplots display pairwise
measures related to \emph{all} dimensions. \emph{Deviation plots},
a new technique, enables grouping and supports a comparison of
measures for a subgroup to the whole cohort.

The possibilities of VA tools can be summarized as follows:
\begin{itemize}
  \item Manual/automatic definition (brushing) of interesting
      parameters and ranges of values in attribute views,
  \item Linking of attribute views for identifying relations,
  \item Analysis across aggregation levels, parameters and
      subjects,
  \item Definition of groups either interactively by means of
      (complex) brushes, or semi-automatically by means of
      clustering, and
  \item Visual queries and direct feedback enable easy
      exploration
\end{itemize}

While having different applications in mind, the Polaris system of
Stolte and colleagues also employs multiple coordinated views to
validate hypotheses \cite{Stolte:2002s}. It uses a variety of
different information visualization techniques to map
ordinal/nominal or quantitative data of a relational data base.
The system itself formulates data base queries and the mapping to
create visualizations for the requested attributes. They choose
the visualization mapping automatically based on the attribute
types that are viewed in context with each other. This allows for
a fast visualization of different attribute combinations in order
to drill down to the information of interest.
General visual analytics tools, such as Polaris and Weaver, in
principle support some of the requirements for epidemiology. The
use of coordinated views, brushes and switches is advantageous
\cite{Weaver:2010ur}. However, they are not designed to cope with
the special requirements of cohort study data and do not directly
support epidemiological workflows. In particular, no support for a
combined assessment of image data and other epidemiological data
is available.\newline

\textbf{Commercial Visual Analytics Systems.} There are a number
of commercial software tools specialized on data visualization. As
an example, Tableau \cite{Mackinlay:2007jp} provides an interface
for creating different visualizations based on attribute
drag-and-drop.\footnote{\url{http://www.tableausoftware.com/},
accessed: 1/31/2014} These approaches deliver fast results with
respect to visualizing different attributes. However, it is not
supported to derive new data, such as scores for diseases, body
mass index and other data that is relevant in epidemiology.
QLikView is a similar tool for visualizing data associations. The
user can design a frontend where associations can be assessed
using multiple information visualizations. Thus, the user can
drill down to the desired information. Statistics features
regarding epidemiological key figures are limited.

Spotfire/IVEE \cite{Ahlberg:1996ep} is able to handle more complex
analysis of data sets and allows for interactive filtering of
attributes. It can be linked to the statistical computing
programming language \texttt{R}, which makes it versatile in
comparison with its competitors. However, users need to be
familiar with the \texttt{R} syntax.

Commercial systems cannot be enhanced or embedded in another
system with hassle-free data exchange. The focus of commercial
data visualization tools is business intelligence yielding a focus
on quantitative data sources. At the same time they excel at
incorporating user collaboration by including comment sections and
share filters or entire setups of a dashboard.

\section{Analysis of Medical Image Data for Epidemiology}
\label{imageDataInEpidemiology}

Medical images are not by themselves useful for epidemiological
analysis, since the semantics of image elements (pixel, voxels) is
too low. The resulting extremely high-dimensional feature space
would be unsuitable for visual analysis. Hence, image data is
sequentially aggregated and reduced. The different steps of this
process i.e. image analysis, shape analysis of extracted objects
and subsequent clustering are characterized and discussed in this
section. Throughout the section, we often refer to one case study,
where MRI data from the lumbar spine is analyzed. The
representation of assumptions on the lumbar spine shape and
location as well as the object detection scheme used are examples
of viable and common approaches. We do not claim that these
techniques are better than any other approaches.

\subsection{Medical Image Analysis}
\label{subsec:imageAnalysis} One of the major purposes of image
analysis for a cohort study is to quantify anatomy, e.g., by
volume, shape or spatial relations between structures.
Quantification may be used to establish a range of \emph{normal
values} for different age groups and to characterize variations.
Such variations may also confirm a disease and thus add to data
derived from clinical tests. Thus, the use of image data enables
more reliable conclusions w.r.t. the incidence and prevalence of
diseases. As discussed in Sect.~\ref{subsec:imageData}, MRI is
particularly interesting because of the wide range of different
information represented in these data. Thus, the examples
discussed in the following relate to MRI data although the
principles are more general.

Due to the wide range of image analysis tasks, the techniques
should be adaptable to different analysis goals. The
parameterization of an image analysis technique should be
intuitive and the interaction should be kept to a minimum. The
latter aspect is particularly important, since often several
thousand datasets need to be analyzed. While largely manual
approaches are acceptable in some clinical settings, such as
radiation treatment planning, they are not feasible in the
evaluation of cohort study data. The reduction of interactivity is
not only a matter of effort, but also to meet the essential goal
of \emph{reproducible approaches}.\newline

\textbf{Detection and segmentation of anatomical structures.} A
modular system is a possible means to meet the central requirement
of image analysis in cohort study data. An example of a cohort
study is the liver segmentation of \cite{Gloger2010}, where
concurrent detection and localization processes are combined for
initial segmentation that is then fine-tuned in a model-driven
segmentation step and finalized by a data-driven correction
process. It has been shown that processes can be re-used and
re-combined to solve a different segmentation task on similar data
(kidney segmentation in MRI \cite{Gloger2012}). Alternatively, the
necessary \emph{domain knowledge}, related to expected size, basic
shape, position and grey values, can be separated from the
detection and segmentation module. This strategy is attractive,
since the user has not to care about the detection process when
changing the application. Two problems have to be addressed in
this case:
\begin{itemize}
  \item What is the expectation about the data support
      integrated to fit a model?
  \item How is the with-class variation of the object in
      search separated from the between-class variation?
\end{itemize}

Point distribution models (PDM) \cite{McInerney1996} address the
second question by training on sample segmentations. Model fitting
is realized by a registration step. When training is not feasible,
a prototypical model may be used instead. It is associated with
restricted input about variation (a few parameters only) and
qualitative knowledge about configuration or part-relationship.

In the following, we describe the linear elastic deformation of a
finite element model (FEM) as a common method to model shape
variation. The user specifies the average shape and two elasticity
parameters: Young's modulus defines how much external force is
needed for a deformation and Poisson's ratio describes how the
deformation is transferred orthogonal to the direction of an
incident force \cite{Rak2013}. The decomposition of the
prototypical shape into finite elements bounded by nodes and
specification of the elasticity parameters results in a stiffness
matrix $K$ that relates the node displacement $u$ to incident
forces $f$ (Eq.~\ref{eq:elasticity}):

\begin{equation}\label{eq:elasticity}
Ku = f
\end{equation}

Different kinds of nodes may be specified that are attracted by
different kinds of forces. Boundary nodes are attracted by the
intensity gradient and inner nodes are attracted from expected
intensity or texture. For letting an FEM move and deform into an
object in an image, deformation is made dependent on time $t$.
Behavior then also depends on mass $M$ of the FEM and
object-specific damping $D$ (Eq.~\ref{eq:damping}).

\begin{equation}\label{eq:damping}
M\ddot{u}(t)+D\dot{u}(t)+Ku(t)=f(t)
\end{equation}

$M$ represents the resistance of the moving FEM to external forces
and allows the model to move over spurious image detail (e.g.,
gradients caused by noise). Damping $D$ avoids oscillation of the
FEM. The system of differential equations is decoupled by solving
the following generalized eigenproblem.

\begin{align*}\label{eq:eigenproblem}
KE=ME\Lambda \text{ with } E^{T}KE=\Lambda \text{ and } E^{T}ME=I
\end{align*}

where $\Lambda$ is the diagonal matrix with real-valued
eigenvalues and $I$ is the identity matrix.


\begin{figure}[!htb]
\begin{center}
    \includegraphics[width=8cm]{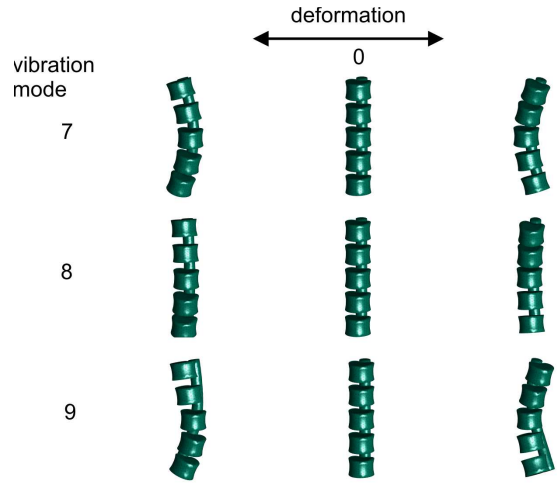}
\end{center}
    \caption{Vibration modes 7 to 9 of a lower spine model. Vibration modes 1 to 6 represent rigid transformations (Courtesy of Marko Rak, University of Magdeburg).}
    \label{fig:image_Anal_Fig1}
\end{figure}

After projecting the data on the eigenvector matrix, the
differential equations can be solved fast and in a stable manner.
Moreover, the projection on the eigenvectors (called \emph{modes
of vibration}, see Fig.~\ref{fig:image_Anal_Fig1}) separates
deformation into components representing rigid transformation,
major deformation modes and remaining minor deformation modes. The
vibration modes can be used similar to the variation modes of an
ASM to derive a quality-of-fit formulation for a fitted model
instance. Since only a few anatomical objects have such a specific
shape that it can be described by a simple deformation model and
since training of additional information should be avoided, it is
useful to complement simple deformation with pre-specified
information on part-relationships. Part-relationships may describe
the configuration of the object of interest w.r.t neighboring
structures or may represent decomposition into parts (see
\cite{Petyt1998}).

Extending the FEM to a hierarchical model requires the
introduction of a second layer FEM. Each sub-shape is represented
by an FEM on the first layer and sub-shape FEMs are connected to
the second layer. The type of connection regulates dependencies
between sub-shapes and may range from distance constraints to
co-deformation. FEMs for the first and second layer are created
and assembled in the same fashion than elements are assembled for
the sub-shape FEM \cite{Rak2013}.\newline

\textbf{Case study: Analysis of vertebrae.} Back pain and related
diseases exhibit a high prevalence and are thus a focus of the
SHIP (recall \cite{Volzke2011}). Specific goals are
\begin{itemize}
\item to define the prevalence of degenerative changes of the
    spine,
\item to identify risk factors for these changes,
\item to correlate degenerative changes with actual symptoms,
    and
\item to better understand the progress from minor disease to
    a severe problem that requires medical treatment.
\end{itemize}

Epidemiologists hypothesize that smoking, heavy physical activity
and a number of drugs that are frequently used are risk factors
for back pain. Based on clinical observations, epidemiologists
suggested to focus on the lumbar spine -- the lowest part of the
spine comprising five vertebrae. As a first step, the spine and
lumbar vertebrae should be detected in T1-weighted and T2-weighted
MRI data from SHIP.

Although local optimization could be complemented by stochastic
global optimization \cite{Engel2010}, Rak and colleagues used only
local optimization, since the initialization is simple for the
given data. The user places a model instance in a sagittal view on
the middle slice of the image sequence which is then transformed
based on local image attributes. The model is constructed
according to the appearance of vertebrae and spine in a sample
image sequence. Vertebra sub-shapes were connected with a spine
sub-shape by a structural model on the second level.

\begin{figure}[!htb]
\begin{center}
\includegraphics[width=10.5cm]{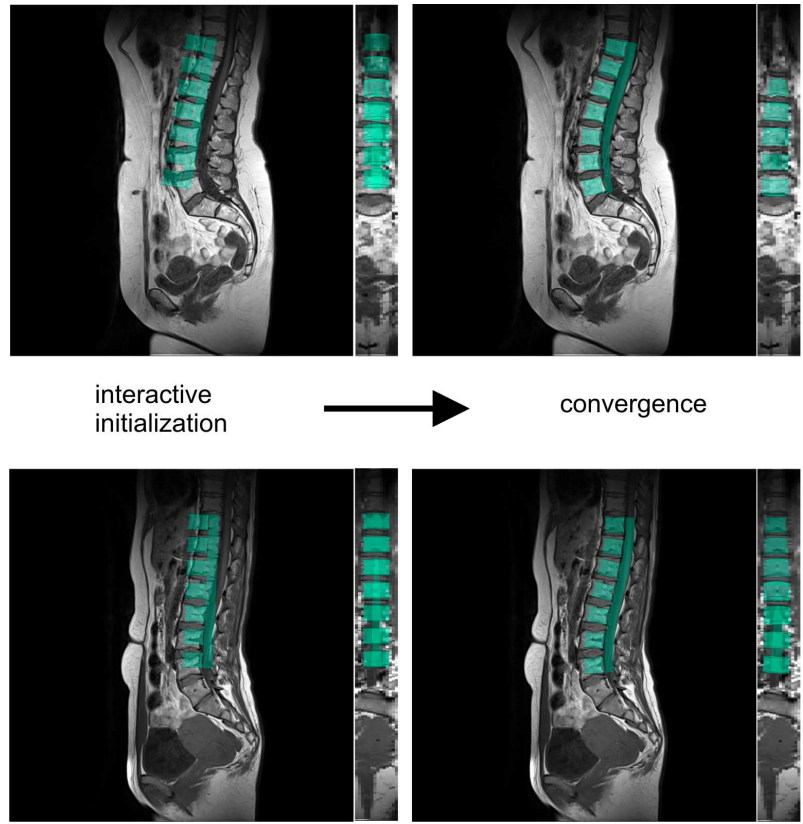}
\end{center}
\caption{Examples for initialization and convergence of model
instances applied to the MRI data (Courtesy of Marko Rak,
University of Magdeburg).} \label{fig:image_Anal_Fig4}
\end{figure}

The spine model supported proper localization of the vertebrae.
Since its most discriminate aspect was the cylindrical shape, it
was represented by a deformable cylinder consisting of inner nodes
only. The vertebrae shape was represented indirectly by inner
nodes as well, since reliability of the intensity gradient was
low. For each of the two shape models, the vertebra and the spine,
a weighted combination of the T1-weighted and the T2-weighted
image was computed as appearance input. Weights for each of the
two models were determined a priori and produced a clearly
recognizable local minimum for vertebra and spine appearance,
respectively. The user placed a model instance in the vicinity of
the object on a sagittal slice. Computation time until convergence
was between 1.1 and 2.6~seconds per case. The method was evaluated
on 49 data sets from the SHIP. The detection was considered
successful if the center of each  vertebra sub-shape was in the
corresponding vertebra in the image data, which was achieved in 48
of the 49 cases (see Fig.~\ref{fig:image_Anal_Fig4} for examples
and \cite{Rak2013} for further details).

\subsection{Shape Analysis}

Epidemiologists are used to work with numerical and categorical
data which then is tested for statistic validity. Medical image
data also allows to consider characteristic object shapes. As an
example, the shape of the liver may depend in a characteristic
manner on infections (hepatitis), alcohol consumption, or obesity.
Eventually, shape characteristics may change even before a disease
becomes symptomatic. If this turns out, shape changes may be
employed as an early stage indicator.

While the quantitative analysis of shapes or parts thereof
(\emph{morphometrics}) is a recent trend in epidemiology due to
the availability of image data, it is established in anatomy and
evolutionary biology.

Shape analysis requires that different shapes are transformed in a
common space, typically by a rigid transformation (translation and
rotation). The parameters of this rigid transformation are
determined in an optimization process that minimizes the distances
of corresponding points. A major challenge is to determine these
corresponding points that serve as landmarks. In particular for
soft tissue structures there are not sufficient recognizable
landmarks and therefore a parameterization is necessary to define
these points. Without going into detail, we assume that this
process is applied to many individual shapes $S_i$, say livers in
a cohort study. Then, for each $S_i$ an optimal non-rigid
transformation to a reference shape $R$ defines a deformation with
displacement vectors for each landmark.

For use in epidemiology, a large set of displacement vectors is
not the right level of granularity. Instead, a few dimensions are
desirable that characterize major differences. Thus, typically a
dimension reduction technique, such as PCA, is employed to
characterize the directions that represent the major differences.
This process may be adapted to specific analysis questions by
assigning individual weights to the landmarks expressing a strong
interest in particular displacements \cite{Hermann2011}. Thus,
epidemiological hypotheses may be incorporated.

While the establishment of point correspondences is often a major
challenge, recently alternative approaches were developed. The
GAMES algorithm (Growing and adaptive meshes) \cite{Ferrarini2007}
creates a data structure to represent the shape variance if no
pairwise correspondence between points is given. However, it can
be prone to errors since it requires a prior registration of
segmentation masks.

Shape analysis, of course, may also be supported by appropriate
visualizations that enable pairwise comparisons and emphasize
differences. In this vein, \cite{Busking2010} presented a system
for shape space exploration based on carefully designed multiple
coordinated views.

\subsection{Analysis of Lumbar Spine Canal Variability}
\label{subsec:AnalysisLSCV}

In Sect.~\ref{subsec:imageAnalysis}, we introduced a case study
related to the analysis of the lumbar spine in cohort data and
explained how the spine and the vertebrae are detected in MRI data
from the SHIP. Here, we extend this discussion by the analysis of
the spinal canal and non-imaging attributes related to back pain.
In the SHIP, attributes related to back pain history, e.g.,
working habits, physical activities, size and weight, are
available to identify and analyze potential correlations with
findings from the MRI data. After careful discussions, we selected
77 attributes (60 are ordinal or nominal and 17 scalar) to
investigate back pain \cite{Klemm_2014_BVM}. Ordinal data are
primarily results of multiple choice questions. The
epidemiologists suggested to focus on the overall shape and
curvature of the spine in that region instead of individual
vertebrae. This overall shape is well characterized by the lumbar
canal. Thus, correlations between the shape of the lumbar spine,
attributes of back pain history and activities both in leisure and
working time may be analyzed.

    \begin{figure}[!htb]
    \sidecaption
    \includegraphics[scale=.193]{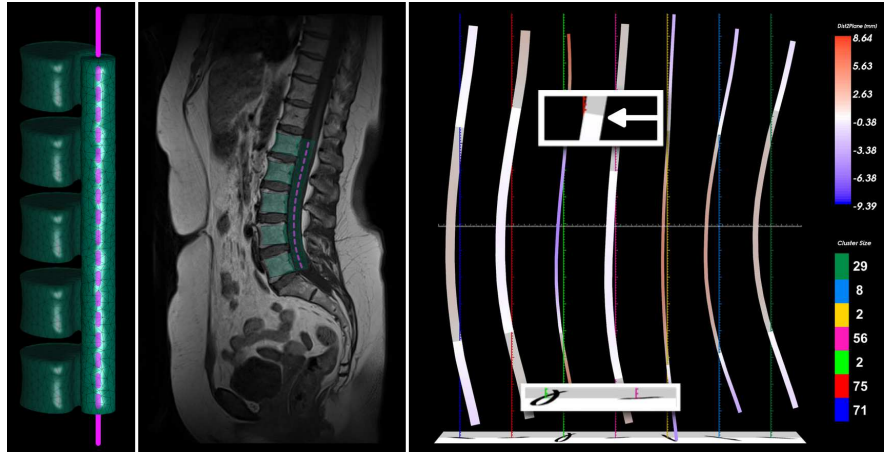}
    \caption{Lumbar spine visualization of 243 female subjects.
        \emph{Left:} Tetrahedron-based finite element model from Rak and colleagues \cite{Rak2013}.
        The dashed purple line indicates the lumbar spine canal centerline
        \emph{Middle:} Model used to detect lumbar spine canal in an MRI scan
        \emph{Right:} Agglomerative hierarchical clustering of 243 centerlines yields seven clusters.
        Their representatives are visualized as ribbons mapping cluster size to width.
        The ribbon color encodes the distance to the semi-transparent plane orthogonal to the view direction
                (lower inset).
        Shadow projections (upper inset) provide an additional visual hint on the curvature extent \cite{Klemm2013}.}
    \label{fig:spineCanalVis}
    \end{figure}

Klemm and colleagues \cite{Klemm2013} extracted, clustered and
visualized spine canal centerlines. Image segmentation of 493~MRI
data sets was carried out automatically using tetrahedron-based
finite element models of vertebrae and spinal canal
\cite{Rak2013}. Using barycentric coordinates of the tetrahedrons,
a centerline consisting of 93 discrete points was extracted for
each segmentation, as seen in Figure~\ref{fig:spineCanalVis}
(left, middle). They served as input for an agglomerative
hierarchical clustering which created groups of subjects based on
differences in shape. This special clustering technique was
chosen, since it produces meaningful results in the clustering of
similar structures, such as fiber tracts derived from MR-Diffusion
Tensor Imaging data \cite{Klemm2013}.

The cluster visualization in Figure~\ref{fig:spineCanalVis}
(right) displays each cluster representative as ribbon in a
sagittal plane. The representative is the centerline with the
smallest sum of distances to all other centerlines, i.e., the
centroid line of the cluster. The width of the ribbon encodes the
cluster count and the color encodes the distance to the sagittal
plane. Shadow projections also (redundantly) convey the distance
to the sagittal plane. This allows to assess the 3D shape in a 2D
projection. The results of the clustering can be used in different
ways.
\begin{itemize}
    \item \textbf{Outlier detection}: Extraordinary shapes
        yield clusters of small size that differ strongly from
        the global mean shape. This can point to pathologies
        or errors in the segmentation process.
    \item \textbf{Hypothesis generation}: Shape groups serve
        as a starting point for an exploratory analysis to
        analyze disease-related correlations. The usual
        workflow requires epidemiologists to define groups,
        e.g., age ranges. Groups calculated solely using shape
        information can be analyzed to detect statistically
        relevant associations in other expositions which can
        lead to new hypotheses.
    \item \textbf{Hypothesis validation}: Clustering based on
        non-image related features can be used to analyze if
        these clusters are correlated to characteristic
        shapes, e.g., a strongly bent spine canal
        representative.
\end{itemize}

Calculating curvature on groups created according to body height
starting from 150~cm in 10~cm steps was performed. Klemm and
colleagues found that taller people have a more straight spine
compared to small people. They also found multiple clusters of
people 10 years above average age across all groups that exhibit a
strong "S" shape of the spine, which was the starting point for
new investigations using expert chosen spine-related attributes.
This method was extended to integrate the relevant information for
identifying correlations. Thus, for a selected cluster,
information related to the distribution of attributes, such as the
back pain history (frequency and intensity of back pain), may be
displayed as a tool tip. The initial observations show, that a box
plot summarizing the distribution is more suitable than the full
histogram.
For routine use in epidemiology, the lumbar spine visualization
(Fig.~\ref{fig:spineCanalVis}) has to be complemented with at
least simple statistics to answer questions, such as: Is there a
statistically significant difference between the curvature of the
lumbar spine canal and back pain frequency? If so, what is the
effect size?

\subsection{Cluster Analysis and Information Space Reduction}

A crucial task in epidemiology is the definition of groups of
subjects. Differences and similarities among groups are
investigated and control groups are defined to detect and assess
the impact and interaction of risk factors to define the relative
risk. A straightforward approach is the manual definition based on
study variables and ranges of interest. A data-driven extension is
the automatic detection of potentially relevant subgroups in the
often high-dimensional data by means of clustering algorithms
\cite{Klemm_2014_BVM}. In particular, the generation of new
hypotheses, which may be tightly connected to the identification
of new groups, benefits from the latter. In clustering, subjects,
being similar with respect to a certain similarity metric, are
grouped in clusters with a low intra-cluster and a high
inter-cluster variance. Particular challenges in the cluster
analysis of cohort study data are:

\begin{itemize}
    \item Missing data, e.g., denied answers to inconvenient questions \cite{Donders2006}
    \item Mixture of scala and categorical study variables
        \cite{AhmadD07}
    \item Time-varying variables in longitudinal studies \cite{Genolini2010}
\end{itemize}

\begin{figure}[!htb]
\begin{center}
\includegraphics[width=11.3cm]{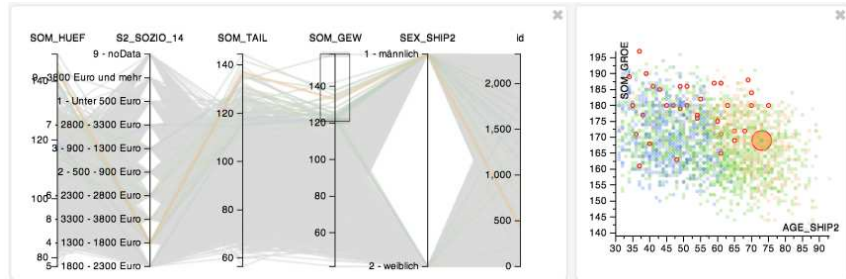}
\end{center}
\caption{A parallel coordinate view enables the selection of a relevant subset (here persons with a weight larger than 120 kg). The scatterplots
represent correlations between age, size and weight. The elements are color-coded according to a clustering result that
yields three clusters. The encircled elements correspond to the selection in the parallel coordinates view.}
\label{fig:toolPaul}
\end{figure}

As a consequence of the first problem, it should be reported to
the user how many datasets were actually used for clustering.
Depending on the chosen attributes, this may be a subset of the
overall amount of data. Incomplete datasets may and should be used
when the relevant data is available. It is essential to use
similarity metrics that consider also ordinal or categorical data.
Usually, the following convention is used: The distance between
datasets equals 1 if their categorical data is different and 0 if
it is the same. With ordinal data, more care is necessary. The
difference between "strongly agree" and "strongly disagree" on a
Likert scale is larger than the difference between "agree" and
"disagree". However, the precise quantification is not
straightforward. As a first step to explore the SHIP data, a
parallel coordinate view is combined with scatterplots and
clustering (Fig.~\ref{fig:toolPaul}).

With respect to clustering algorithms, it is essential that the
number of resulting groups has not to be specified in advance.
Moreover, algorithms are preferred that allow outliers instead of
forcing all elements to be part of a cluster. Outliers may be
particularly interesting and thus serve as a starting point for
further investigation. Of course, they may also indicate a bad
quality of some data. In our experiments, density-based clustering
with DBSCAN \cite{Ester:1996} produced plausible results when
applied to non-image data of the SHIP study. The DBSCAN result is
sensitive to the \emph{minPoints} parameter that determines the
minimum size of a cluster. Some cycles of clustering are necessary
to make a suitable choice. In this process, an appropriate
visualization is essential to easily understand the results. A
visualization that conveys the location, size and shape of
clusters is difficult in case that clustering is applied to more
than two-dimensional data. A recently developed approach enables
3D visualizations of clustering results with very low levels of
occlusion \cite{Glasser_2014_GRAPP}.

The majority of cohort-studies are restricted to non-image data,
i.e., categorical and scalar data, which may directly serve as
input for a clustering algorithm. In \cite{Axen2011}, 176 patients
with lower back pain have been monitored over six months via text
messages describing their bothersomeness. All patients received
chiropractic treatment. A hierarchical clustering of the
individual temporal courses of bothersomeness revealed groups of
patients who responded differently to the therapy, which may
improve the optimal individual treatment selection. Hypotheses
about the differences between paternal age-related schizophrenia
(PARS) and other cases of schizophrenia were generated in
\cite{Lee2011}. A k-means clustering of demographic variables,
symptoms, cognitive tests and olfaction for 136 subjects (34 with
PARS) delivered clusters containing a high concentration of PARS
cases. Significant characteristics of these clusters may give a
hint on features of PARS improving its dissociation of other cases
of schizophrenia.

In analyzing image-centric cohort study data, besides categorical
and scalar data, images may be clustered for group definition.
Image intensities, segmentation results, e.g., the surface of the
segmented liver or the centerline of the spinal canal (recall
Section~\ref{subsec:AnalysisLSCV}) and derived information, e.g.,
liver tissue texture, liver volume and spinal canal centerline
geometry, may serve as input. In \cite{Seiler2012}, the anatomical
variation of the mandibles is assessed across a population. For
treating mandible fractures, subjects with similar characteristics
are grouped in clusters and a suitable implant is designed per
cluster. The clustering algorithm k-means is applied to
transformation parameters of a locally affine registration between
all mandible surfaces segmented in CT data. A cohort of 50
patients with suspicious breast lesions was investigated by means
of dynamic contrast enhanced MRI (DCE-MRI) in \cite{Preim2012}.
Each lesion was clustered according to its perfusion
characteristics by means of a region merging approach. Perfusion
is represented by the temporal course of the DCE-MRI signal
intensities. The clustering itself did not generate groups of
patients here, but based on each individual number of clusters and
their perfusion characteristics, two groups of lesions could be
defined: benign and malignant. These groups were then compared to
histological results from core needle biopsy.

Investigating high-dimensional non-image cohort study data often
benefits from an information space reduction, e.g., by means of
PCA. Plotting the data for inspection in a lower dimensional space
while capturing the greatest level of variation, e.g., a
scatterplot of the first two principal components,  as well as
detecting trends in the data and ordering them according to the
variance they describe are important applications of PCA. In
\cite{Robertson2008}, symptom data gathered in interviews of 410
people with Turret syndrome was investigated in order to specify
homogeneous symptom categories for a better characterization of
the disease's phenotype. First, clusters of symptom variables were
generated using \emph{agglomerative hierarchical clustering}.
Then, for each cluster and each participant a score was computed
equal to the sum of present symptom variables in the cluster.
These scores were the input for PCA, which produced homogeneous
symptom categories, sorted according to their percentage of
represented symptomatic variance. In \cite{Steenwijk:2010wo},
scatterplots of variables from cohort study data are extended by
superimposing PCA ellipses. The ellipses are computed per group of
subjects and illustrate its global distribution with respect to
the two opposed variables. They are spanned by the principal
component axes of a groups data points and centered at their mean.
Optionally, their transparency is adjusted with respect to the
groups confidence, i.e. the number of contained subjects.

\subsection{Categorical Data}
Cohort study data sets comprise many categorical data such as
answers to questionnaires or categorizations that may result from
binning continuous data, such as intervals of income or age
groups. These data are \emph{discrete} and exhibit a \emph{low
range}. While data resulting from binning or from answers marked
at a Likert scale have an inherent order (ordinal data), often no
inherent order exists. Standard information visualization methods
like scatterplots and parallel coordinates are designed for
continuous data and thus not ideal for displaying categorical
data, since many data points occlude each other.

\begin{figure}[!htb]
\begin{center}
\includegraphics[width=10cm]{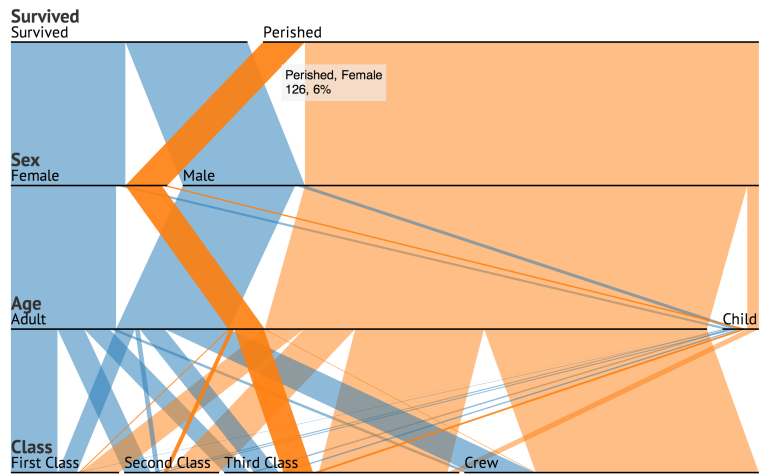}
\end{center}
\caption{Parallel sets are designed to explore categorical data.
Parallel sets are based on the parallel coordinates layout, but it
maps frequency of the data instead of rendering them just as data
points. The user can interactively remap the data to new
categorizations as well as highlight entries to examine their
distribution along other mapped dimensions. The displayed data set
shows the distribution of passengers of the RMS Titanic and
whether they survived the sinking of the ship \cite{Davies2013}.}
\label{fig:ParallelSet}
\end{figure}
Categorical data are often visualized using boxes where the width
is scaled to frequency \cite{Wittenburg:2001}.
These approaches use much space and are also not well suited for
encoding multidimensional relationships. \emph{Parallel sets},
introduced by Bendix and colleagues, comprise the same layout as
parallel coordinates, "but the continuous axes were replaced with
sets of boxes \ldots scaled to the frequency of the category"
\cite{Bendix:2005vf} (see Fig. \ref{fig:ParallelSet}).
Selecting an attribute will map each category to a distinct color
so that they can be traced through all visualized dimensions.
Highlighting a category may be realized by drawing the selected
category in a higher saturation leading to a pop-out effect. It is
also useful to display a histogram for the selected category
annotated with statistical information. Selecting a category will
only display the particular box on an axis and make more room for
connections to other axes.
This is especially helpful if the number of categories for one
dimension is large. In case of data without inherent order,
reordering of axes is possible and supports the exploration by
providing more comprehensible layouts.

\section{Concluding Remarks}

A variety of large prospective cohort studies are established and
ongoing. They generate a wealth of potentially relevant
information for assessing risks, for estimating costs involved in
treatment and thus inform health policy makers with respect to
potentially preventive measures or cost-limiting initiatives.
Despite the great potential of data mining and interactive
visualization, none of these studies included such activities in
their original planning. The role of computer science, so far, was
limited to database management and data security. Visual analytics
has a great potential for exploring complex health-related data,
as recently shown by Zhang and colleagues \cite{Zhang:2013} for
clinical applications, such as treatment planning. Similar
techniques may be employed to address epidemiology research. In
contrast to clinical applications, where a severe time pressure
leads to strongly guided workflows, epidemiology research benefits
from powerful and flexible tools that enable and support
\emph{exploration}. Currently, existing and widespread information
visualization and analytics techniques are employed and adapted to
epidemiologic data. The high-dimensional nature of these data,
however, also requires to develop new techniques.

The specific and new aspect discussed in this paper was the
integration of image data, information derived from image data,
such as spine curvature-related measures, and more traditional
socio-demographics data.\newline

\textbf{Future work.} So far, visual analytics research and
software was rarely focused on epidemiology. Thus, to adapt visual
analytics to epidemiology and to integrate the solutions with
tools familiar to epidemiologists is necessary. In epidemiology,
national studies are prevailing, which is due to the large amount
of legislative conditions to be considered. International studies
would enable to explore diseases with lower prevalence, subtypes
of diseases that occur rarely, e.g., cancer in early age, and
specific questions of spatial epidemiology. The SHIP Brazil study
will provide such information. It was recently initiated to
perform a study in Brazil according to the standards and
experiences gained in the German SHIP study. A nasty but essential
problem of image-based epidemiologic study is quality control.
Research efforts are necessary to automatically check whether
image data fully cover the target region, whether the alignment of
slices, e.g., in heart imaging, is correct and whether severe
artifacts appear, e.g., motion artifacts. This chapter discussed
exciting developments related to the combined use of radiologic
image and more classical epidemiological data. The next wave is
already clearly recognizable: genetics information will be
integrated in the search for early markers associated with risks
for diseases.\newline

\begin{acknowledgement} We want to thank Lisa Fraunstein, David Kilias and David
Perlich who supported our analysis of the SHIP data as student
workers as well as Marko Rak who provided the detection algorithm
for the vertabrae and Myra Spilopoulou for fruitful discussions on
clustering and data mining (all University of Magdeburg). We thank
Martijn Steenwijk for providing images from his work and Charl
Botha for fruitful discussions. Matthias G{\"u}nther (Fraunhofer
MEVIS) explained us quality aspects of MR imaging in epidemiologic
studies. This work was supported by the DFG Priority Program 1335:
Scalable Visual Analytics. SHIP is part of the Community Medicine
Research net of the University of Greifswald, Germany, which is
funded by the Federal Ministry of Education and Research (grant
no. 03ZIK012), the Ministry of Cultural Affairs as well as the
Social Ministry of the Federal State of Mecklenburg-West
Pomerania.

\end{acknowledgement}

\bibliographystyle{styles/spmpsci}
\bibliography{bibliography}

\begin{thebibliography}{10}
\providecommand{\url}[1]{{#1}}
\providecommand{\urlprefix}{URL }
\expandafter\ifx\csname urlstyle\endcsname\relax
  \providecommand{\doi}[1]{DOI~\discretionary{}{}{}#1}\else
  \providecommand{\doi}{DOI~\discretionary{}{}{}\begingroup
  \urlstyle{rm}\Url}\fi

\bibitem{Donders2006}
{Review: A Gentle Introduction to Imputation of Missing Values}.
\newblock Journal of Clinical Epidemiology \textbf{59}(10), 1087 -- 1091 (2006)

\bibitem{Ahlberg:1996ep}
Ahlberg, C.: {Spotfire: an information exploration environment}.
\newblock SIGMOD Record \textbf{25}(4) (1996)

\bibitem{AhmadD07}
Ahmad, A., Dey, L.: {A k-mean Clustering Algorithm for Mixed Numeric and
  Categorical Data}.
\newblock Data Knowl. Eng. \textbf{63}(2), 503--527 (2007)

\bibitem{Axen2011}
Ax{\'e}n, I., Bodin, L., Bergstr{\"o}m, G., Halasz, L., Lange, F., L{\"o}vgren,
  P.W., Rosenbaum, A., Leboeuf-Yde, C., Jensen, I.: {Clustering Patients on the
  Basis of Their Individual Course of Low Back Pain Over a Six Month Period}.
\newblock B M C Musculoskeletal Disorders \textbf{12}, 99--108 (2011)

\bibitem{Beale:2008dz}
Beale, L.L., Abellan, J.J., Hodgson, S.S., Jarup, L.L.: {Methodologic issues
  and approaches to spatial epidemiology}.
\newblock Environmental health perspectives \textbf{116}(8), 1105--1110 (2008)

\bibitem{Bendix:2005vf}
Bendix, F., Kosara, R., Hauser, H.: {Parallel sets: Visual analysis of
  categorical data} pp. 133--140 (2005)

\bibitem{Blaas2007}
Blaas, J., Botha, C.P., Post, F.H.: {Interactive Visualization of Multi-Field
  Medical Data Using Linked Physical and Feature-Space Views}.
\newblock In: Proc. of EuroVis, pp. 123--130 (2007)

\bibitem{Busking2010}
Busking, S., Botha, C.P., Post, F.H.: {Dynamic Multi-View Exploration of Shape
  Spaces}.
\newblock Computer Graphics Forum \textbf{29}(3), 973--982 (2010)

\bibitem{Chui:2011gm}
Chui, K.K., Wenger, J.B., Cohen, S.A., Naumova, E.N.: {Visual analytics for
  epidemiologists: understanding the interactions between age, time, and
  disease with multi-panel graphs}.
\newblock PloS one \textbf{6}(2), e14,683 (2011)

\bibitem{Davies2013}
Davies, J.: {Parallel Set of the Titanic Data Set}.
\newblock \url{http://www.jasondavies.com/parallel-sets/} (2012).
\newblock [Online; accessed 30-January-2014]

\bibitem{Engel2010}
Engel, K., Toennies, K.D.: {Hierarchical vibrations for part-based recognition
  of complex objects}.
\newblock Pattern Recognition \textbf{43}(8), 2681--2691 (2010)

\bibitem{Ester:1996}
Ester, M., Kriegel, H.P., Sander, J., Xu, X.: {A Density-Based Algorithm for
  Discovering Clusters in Large Spatial Databases with Noise}.
\newblock In: Proc. of the Second International Conference on Knowledge
  Discovery and Data Mining (KDD), pp. 226--231 (1996)

\bibitem{Ferrarini2007}
Ferrarini, L., Olofsson, H., Palm, W., Vanbuchem, M., Reiber, J.,
  Admiraalbehloul, F.: {GAMEs: Growing and adaptive meshes for fully automatic
  shape modeling and analysis}.
\newblock Medical Image Analysis \textbf{11}(3), 302--314 (2007)

\bibitem{Fletcher2011}
Fletcher, R.H., Fletcher, S.W.: Clinical Epidemiology.
\newblock Lippincott Williams \& Wilkins (2011)

\bibitem{Genolini2010}
Genolini, C., Falissard, B.: {KmL: k-means for Longitudinal Data}.
\newblock Comput. Stat. \textbf{25}(2), 317--328 (2010)

\bibitem{Glasser_2014_GRAPP}
Gla\"{ss}er, S., Lawonn, K., Preim, B.: {Visualization of 3D Cluster Results
  for Medical Tomographic Image Data}.
\newblock In: {In Proc. of Conference on Computer Graphics Theory and
  Applications (VISIGRAPP/GRAPP)}, pp. 169--176 (2014)

\bibitem{Gloger2010}
Gloger, O., J., K., Stanski, A., V{\"o}lzke, H., Puls, R.: {A fully automatic
  three-step liver segmentation method on LDA-based probability maps for
  multiple contrast MR images}.
\newblock Magnetic Resonance Imaging \textbf{28}(6), 882--897 (2010)

\bibitem{Gloger2012}
Gloger, O., Toennies, K.D., Liebscher, V., Kugelmann, B., Laqua, R., H., V.:
  {Prior shape level set segmentation on multistep generated probability maps
  of MR datasets for fully automatic kidney parenchyma volumetry}.
\newblock IEEE Transactions on Medical Imaging \textbf{31}(2), 312--325 (2012)

\bibitem{Gresh:2000}
Gresh, D.L., Rogowitz, B.E., Winslow, R.L., Scollan, D.F., Yung, C.K.: Weave: a
  system for visually linking 3-d and statistical visualizations, applied to
  cardiac simulation and measurement data.
\newblock In: Proc. of IEEE Visualization, pp. 489--492 (2000)

\bibitem{Hegenscheid2009}
Hegenscheid, K., K{\"u}hn, J.P., V{\"o}lzke, H., Biffar, R., Hosten, N., Puls,
  R.: {Whole-body magnetic resonance imaging of healthy volunteers: pilot study
  results from the population-based SHIP study}.
\newblock {Fortschritte auf dem Gebiet der R"{\o}ntgenstrahlen und der
  bildgebenden Verfahren (R\"{o}fo)} \textbf{181 (8)}, 748--759 (2009)

\bibitem{Hermann2011}
Hermann, M., Schunke, A.C., Klein, R.: Semantically steered visual analysis of
  highly detailed morphometric shape spaces.
\newblock In: Proc. of IEEE Symposium on Biological Data Visualization
  (BioVis), pp. 151--158 (2011)

\bibitem{Hofman2009}
Hofman, A., Breteler, M.M.B., van Duijn, C.M., Janssen, H.L.A., Krestin, G.P.,
  Kuipers, E.J., Stricker, B.H.C., Tiemeier, H., Uitterlinden, A.G.,
  Vingerling, J.R., Witteman, J.C.M.: {The Rotterdam Study: 2010 objectives and
  design update}.
\newblock European Journal of Epidemiology \textbf{24}, 553--572 (2009)

\bibitem{Hofman2011}
Hofman, A., van Duijn, C.M., Franco, O.H., et~al.: {The Rotterdam Study: 2012
  objectives and design update}.
\newblock European Journal of Epidemiology \textbf{26}, 657--686 (2011)

\bibitem{Jerrett:2010}
Jerrett, M., Gale, S., Kontgis, C.: {Spatial Modeling in Environmental and
  Public Health Research}.
\newblock International Journal of Environmental Research in Public Health
  \textbf{7}(16), 1302--1329 (2010)

\bibitem{Klemm_2014_BVM}
Klemm, P., Frauenstein, L., Perlich, D., Hegenscheid, K., V{\"o}lzke, H.,
  Preim, B.: {Clustering Socio-demographic and Medical Attribute Data in Cohort
  Studies}.
\newblock In: Proc. of Bildverarbeitung fuer die Medizin (BVM) (2014)

\bibitem{Klemm2013}
Klemm, P., Lawonn, K., Rak, M., Preim, B., T{\"o}nnies, K., Hegenscheid, K.,
  V{\"o}lzke, H., Oeltze, S.: {Visualization and Analysis of Lumbar Spine Canal
  Variability in Cohort Study Data}.
\newblock In: Proc. of Vision, Modeling, Visualization (VMV), pp. 121--128
  (2013)

\bibitem{Lee2011}
Lee, H., Malaspina, D., Ahn, H., Perrin, M., Opler, M.G., Kleinhaus, K.,
  Harlap, S., Goetz, R., Antonius, D.: {Paternal Age Related Schizophrenia
  (PARS): Latent Subgroups Detected by k-means Clustering Analysis}.
\newblock Schizophrenia Research \textbf{128}(1–3), 143 -- 149 (2011)

\bibitem{Mackinlay:2007jp}
Mackinlay, J., Hanrahan, P., Stolte, C.: {Show me: automatic presentation for
  visual analysis}.
\newblock IEEE Transactions on Visualization and Computer Graphics
  \textbf{13}(6), 1137--1144 (2007)

\bibitem{Marathe2013}
Marathe, M., Vullikanti, A.K.S.: Computational epidemiology.
\newblock Communications of the ACM \textbf{56}(7), 88--96 (2013)

\bibitem{McInerney1996}
McInerney, T., Terzopoulos, D.: {Deformable models in medical image analysis: a
  survey}.
\newblock Medical Image Analysis \textbf{1}(2), 91--108 (1996)

\bibitem{Pearce2006}
Pearce, N., Merletti, F.: {Complexity, simplicity, and epidemiology}.
\newblock International Journal of Epidemiology \textbf{35}(3), 515--519 (2006)

\bibitem{Peterson2013}
Petersen, S.E., Matthews, P.M., Bamberg, F., et~al.: {Imaging in population
  science: cardiovascular magnetic resonance in 100,000 participants of UK
  Biobank - rationale, challenges and approaches}.
\newblock {J. Cardiovasc Magn Reson.} \textbf{28}, 15--46 (2013)

\bibitem{Petyt1998}
Petyt, M.: Introduction to finite element vibration analysis.
\newblock Cambridge University Press (1998)

\bibitem{Preim2012}
Preim, U., Gla{\ss}er, S., Preim, B., Fischbach, F., Ricke, J.: {Computer-aided
  Diagnosis in Breast DCE-MRI-Quantification of the Heterogeneity of Breast
  Lesions}.
\newblock {European Journal of Radiology} \textbf{81}(7), 1532--1538 (2012)

\bibitem{Rak2013}
Rak, M., Engel, K., T\"{o}nnies, K.D.: {Closed-Form Hierarchical Finite Element
  Models for Part-Based Object Detection}.
\newblock In: Proc. of Vision, Modeling, Visualization (VMV), pp. 137--144
  (2013)

\bibitem{Robertson2008}
Robertson, M.M., Althoff, R.R., Hafez, A., Pauls, D.L.: {Principal Components
  Analysis of a Large Cohort With Tourette Syndrome}.
\newblock The British Journal of Psychiatry \textbf{193}(1), 31--36 (2008)

\bibitem{Seiler2012}
Seiler, C., Pennec, X., Reyes, M.: {Capturing the Multiscale Anatomical shape
  Variability with Polyaffine Transformation Trees}.
\newblock Medical Image Analysis \textbf{16}(7), 1371 -- 1384 (2012)

\bibitem{Steenwijk:2010wo}
Steenwijk, M.D., Milles, J., van Buchem, M.A., Reiber, J.H.C., Botha, C.P.:
  {Integrated Visual Analysis for Heterogeneous Datasets in Cohort Studies}.
\newblock In: Proc. of IEEE VisWeek Workshop on Visual Analytics in Health Care
  (2010)

\bibitem{Stolte:2002s}
Stolte, C., Tang, D., Hanrahan, P.: {Polaris: A system for query, analysis, and
  visualization of multidimensional relational databases}.
\newblock IEEE Transactions on Visualization and Computer Graphics
  \textbf{8}(1), 52--65 (2002)

\bibitem{Thew:2009vz}
Thew, S., Sutcliffe, A., Procter, R., de~Bruijn, O., McNaught, J., Venters,
  C.C., Buchan, I.: {Requirements engineering for e-Science: experiences in
  epidemiology}.
\newblock IEEE Software \textbf{26}(1), 80--87 (2009)

\bibitem{Turkay:2013}
Turkay, C., Lundervold, A., Lundervold, A.J., Hauser, H.: Hypothesis generation
  by interactive visual exploration of heterogeneous medical data.
\newblock In: Proc. of Human-Computer Interaction and Knowledge Discovery in
  Complex, Unstructured, Big Data, pp. 1--12 (2013)

\bibitem{Volzke2011}
V{\"o}lzke, H., Alte, D., Schmidt, C., et~al.: {Cohort Profile: The Study of
  Health in Pomerania}.
\newblock International Journal of Epidemiology \textbf{40}(2), 294--307 (2011)

\bibitem{Volzke2005}
V{\"o}lzke, H., Baumeister, S.E., Alte, D., Hoffmann, W., Schwahn, C., Simon,
  P., John, U., Lerch, M.M.: {Independent Risk Factors for Gallstone Formation
  in a Region with High Cholelithiasis Prevalence}.
\newblock Digestion \textbf{71}, 97--105 (2005)

\bibitem{Weaver:2010ur}
Weaver, C.: {Cross-filtered views for multidimensional visual analysis}.
\newblock IEEE Transactions on Visualization and Computer Graphics
  \textbf{16}(2), 192--204 (2010)

\bibitem{Wittenburg:2001}
Wittenburg, K., Lanning, T., Heinrichs, M., Stanton, M.: Parallel bargrams for
  consumer-based information exploration and choice.
\newblock In: Proc. of the ACM symposium on User interface software and
  technology (UIST), pp. 51--60 (2001)

\bibitem{YstadPhD}
Ystad, M.: Quantitative structural and functional brain imaging in cognitive
  aging.
\newblock Ph.D. thesis, University of Bergen (2010)

\bibitem{Ystad2010}
Ystad, M., Eichele, T., Lundervold, A.J., Lundervold, A.: Subcortical
  functional connectivity and verbal episodic memory in healthy
  elderly---resting state fmri study.
\newblock NeuroImage \textbf{52}(1), 379--388 (2010)

\bibitem{Ystad2009}
Ystad, M., Lundervold, A.J., Wehling, E., Espeseth, T., Rootwelt, H., Westlye,
  L., Andersson, M., Adolfsdottir, S., Geitung, J., Fjell, A., Reinvang, I.,
  Lundervold, A.: Hippocampal volumes are important predictors for memory
  function in elderly women.
\newblock BMC Medical Imaging \textbf{9}(1), 1--15 (2009)

\bibitem{Zhang:2012uw}
Zhang, Z., Gotz, D., Perer, A.: {Interactive Visual Patient Cohort Analysis}.
\newblock In: Proc. of IEEE VisWeek Workshop on Visual Analytics in Healthcare
  (2012)

\bibitem{Zhang:2013}
Zhang, Z., Wang, B., Ahmed, F., Ramakrishnan, I., Viccellio, A., Zhao, R.,
  Mueller, K.: {The Five W's for Information Visualization with Application to
  Healthcare Informatics}.
\newblock IEEE Transactions on Visualization and Computer Graphics
  \textbf{19}(11), 379--388 (2013)

\end{thebibliography}
\end{document}